\documentclass{article} %
\usepackage{iclr2026_conference,times}

\usepackage{amsmath,amsfonts,bm}

\def\eqref#1{equation~\ref{#1}}

\def\1{\bm{1}}

\DeclareMathAlphabet{\mathsfit}{\encodingdefault}{\sfdefault}{m}{sl}
\SetMathAlphabet{\mathsfit}{bold}{\encodingdefault}{\sfdefault}{bx}{n}

\usepackage{hyperref}
\usepackage{url}
\usepackage{color,soul}
\usepackage{graphicx}

\title{Learning to Staff: Offline Reinforcement Learning and Fine-Tuned LLMs for Warehouse Staffing Optimization}

\author{Kalle Kujanp\"a\"a, Yuying Zhu, Kristina Klinkner \& Shervin Malmasi \\
Amazon, Fulfillment Technologies \& Robotics \\
\texttt{\{kujanpaa,imyuying,klinkner,malmasi\}@amazon.com} \\
}

\iclrfinalcopy %
\begin{document}

\maketitle

\begin{abstract}
We investigate machine learning approaches for optimizing real-time staffing decisions in semi-automated warehouse sortation systems. Operational decision-making can be supported at different levels of abstraction, with different trade-offs. We evaluate two approaches, each in a matching simulation environment. First, we train custom Transformer-based policies using offline reinforcement learning on detailed historical state representations, achieving a 2.4\% throughput improvement over historical baselines in learned simulators. In high-volume warehouse operations, improvements of this size translate to significant savings. Second, we explore LLMs operating on abstracted, human-readable state descriptions. These are a natural fit for decisions that warehouse managers make using high-level operational summaries. We systematically compare prompting techniques, automatic prompt optimization, and fine-tuning strategies. While prompting alone proves insufficient, supervised fine-tuning combined with Direct Preference Optimization on simulator-generated preferences achieves performance that matches or slightly exceeds historical baselines in a hand-crafted simulator. Our findings demonstrate that both approaches offer viable paths toward AI-assisted operational decision-making. Offline RL excels with task-specific architectures. LLMs support human-readable inputs and can be combined with an iterative feedback loop that can incorporate manager preferences.

\end{abstract}

\section{Introduction}

In a semi-automated sortation system, warehouse managers make staffing decisions by monitoring throughput rates, buffer levels, and queue lengths across processing lines. They aim to keep the system balanced, and they reassign workers to address emerging bottlenecks. These decisions are made periodically throughout a shift and they have effects across the system: reassigning a worker to clear one bottleneck can lead to another bottleneck emerging elsewhere. Furthermore, while moving a worker might improve throughput in isolation, each reallocation is associated with a cost. The problem is further complicated by dependencies between parallel processing lines. The different stages of these lines interact through shared buffers. Furthermore, the buffer states can have non-linear effects on the throughput and the dynamics cannot easily be expressed in closed form. Because of these factors, staffing allocation is a strategic decision problem. Actions have delayed consequences and local optimization can harm the global performance. Furthermore, managers must understand how their current choices constrain or, alternatively, expand future alternatives. The action space grows exponentially in the number of workers. Hence, modeling this problem analytically is difficult, motivating learned approaches.

Operational decision-making can be supported at different levels of abstraction, and each level has its own trade-offs. When the states are represented in a detailed fashion, we can capture fine-grained dynamics. However, this requires specialized neural network architectures to handle high-dimensional state spaces and produce recommendations that are difficult for humans to interpret or override. In turn, abstracted representations align with how managers reason about operations and make it easier for managers to apply their warehouse-specific knowledge. However, this abstraction can come at a cost, and may lead to inferior recommendation quality. We investigate the requirements of each approach empirically in the context of warehouse staffing allocations. Throughout our work, our objective is to support the human decision-makers in this task, as managers have significant tacit knowledge that is not captured in the state representations of an AI system. We develop systems that generate decision suggestions that the manager can evaluate, modify, or override. In a deployed system, the manager's decisions would be logged, and these could be useful for continual learning. We simulate this interaction loop during development using preference optimization, where short simulated rollouts act as a proxy for manager feedback, which allows us to validate this framework before deployment with human preferences.

LLMs have played many roles in combinatorial optimization, including problem modeling, heuristic design, code generation for solvers, and direct solution generation \citep{da2025large}. In our work, we focus on direct solution generation, and also touch upon heuristic design. In the context of logistics and warehouse operations, previous work has applied LLMs to routing \citep{huang2024words, li2025ars, tranlarge, ieva2025enhancing}. In warehouses, LLMs have been applied to package movement optimization \citep{berlec2025exploring}, warehouse bottleneck identification \citep{parekh2025leveraging}, and delivery scheduling \citep{kmiecik2025creating}, among other tasks. However, so far, staffing has been underexplored as an application area.

In our work, we focus on staffing and study two paradigms at their natural abstraction levels: offline reinforcement learning and large language models. These families of methods operate on fundamentally different state-action representations, which makes direct comparison infeasible. Instead, we investigate whether each approach can learn to optimize sortation system staffing given its natural problem representation to support the human decision-makers in the warehouses. Offline reinforcement learning operates on high-dimensional, detailed state representations. In theory, offline RL allows us to learn from the most effective decisions in the historical data, and potentially allows exceeding the best historical performance in the dataset. We develop a Transformer-based Graph Neural Network (GNN) architecture that models interactions between system components, which enables tractable training and fast inference. Large language models, in turn, operate on abstracted, human-readable descriptions that mirror the operational summaries that managers use. We systematically evaluate the degree of task-specific adaptation that is required, from zero-shot prompting through direct preference optimization. Our iterative preference optimization simulates repeated simulator-manager interactions: the model proposes staffing changes, a simulated manager compares alternatives, and the model learns from this feedback. 

Our contributions are as follows:

\begin{itemize}
\item We develop a Transformer-GNN architecture for offline RL on detailed state representations, achieving a 2.4\% throughput improvement over replayed human decisions in learned simulators.
\item We show that small and mid-sized, open-weight LLMs operating on abstracted, human-readable states require substantial task-specific adaptation: prompting alone proves insufficient, but supervised fine-tuning approaches historical baseline performance.
\item We demonstrate that iterative preference optimization, which simulates manager feedback using a simulator, enables LLMs to match or slightly outperform historical baselines. The framework is designed such that simulator-generated preferences can be replaced with real manager feedback without changing the method, laying groundwork for human-AI collaborative learning with real manager feedback.
\end{itemize}

\section{Problem Setting}

\subsection{Domain Description}

We study real-time staffing allocation in a large-scale sortation system that processes customer orders. The inputs to the system are the individual products ordered by the customers, and the outputs of the system are packages that can be delivered to the customers. The system consists of multiple parallel processing lines. Each of them contains several sequential stages. Workers are assigned to specific stages within lines, and stages are connected by conveyor belts or buffers. A manager makes staffing decisions by reallocating workers from one stage to another, or from one processing line to another. In our setting, we assume that these reallocation decisions are made periodically.

Critically, there are multiple factors that make this decision problem complicated. The parallel processing lines are not fully independent. Rather, they interact through shared buffers and internal system states. For instance, operating adjacent lines simultaneously increases the risk of jammed items or buffer overflows. Consequently, the manager must make decisions at different levels: which lines to operate (opening or closing lines based on demand and system state), and how to staff the active lines. Closing an understaffed line and redistributing its workers to other lines can be preferable to operating many lines at reduced efficiency.

In addition, the system has highly non-linear state-dependent dynamics that make analytical optimization difficult. Buffers between stages lead to feedback loops that affect throughput. If a downstream buffer fills, upstream workers slow down, as they have nowhere to send completed items. Conversely, when a buffer empties, downstream workers slow down as there is nothing to process. These relationships can be non-linear. Productivity does not always decrease linearly with buffer fullness, but there are complex interactions and threshold effects. Decisions that are locally optimal can create bottlenecks that reduce global throughput. Hence, we aim to learn these dynamics from operational data rather than specify them analytically. Finally, each reallocation incurs a productivity cost: workers must leave their current station, physically move to a new location, and adjust to different tasks.

\subsection{Decision Problem}

We formalize staffing allocation as a sequential decision problem. The state of the system consists of the current worker assignment across all lines and stages, current throughput rates at each stage, and internal system status indicators that affect processing capacity. The action space consists of worker assignments, specifying which workers to move to which destinations. This implicitly determines which lines are staffed and active. We work in the episodic reinforcement learning setting, with a typical length of an episode being around 50 steps. We can potentially move multiple workers simultaneously, or maintain the current staffing. The objective is to maximize system throughput over a work shift, which requires balancing immediate bottleneck resolution against longer-term balancing of the system. A reallocation that helps now can be harmful later. We measure the throughput improvement as relative gain in units per hour. The action space is combinatorially huge: tens to hundreds of workers can be reassigned to any valid position, subject to stage-specific capacity constraints.

Our experiments use a dataset of historical sortation system trajectories. The dataset has been split into a disjoint train and test set, where the train set consists of 31,366 work shifts (trajectories), with a total of 501,856 transitions recorded at a 5-minute frequency, corresponding to a little under 42,000 hours of sortation system data across different physical sites, collected over multiple months. The test set size is 8,228 work shifts, corresponding to a total of 131,648 transitions. For each time point, we have the anonymized worker IDs, the roles that they were assigned to, the stage-specific output rates and the buffer states.

\subsection{Evaluation Approach}

\begin{figure}[t]
\begin{center}
\includegraphics[width=0.6\textwidth]{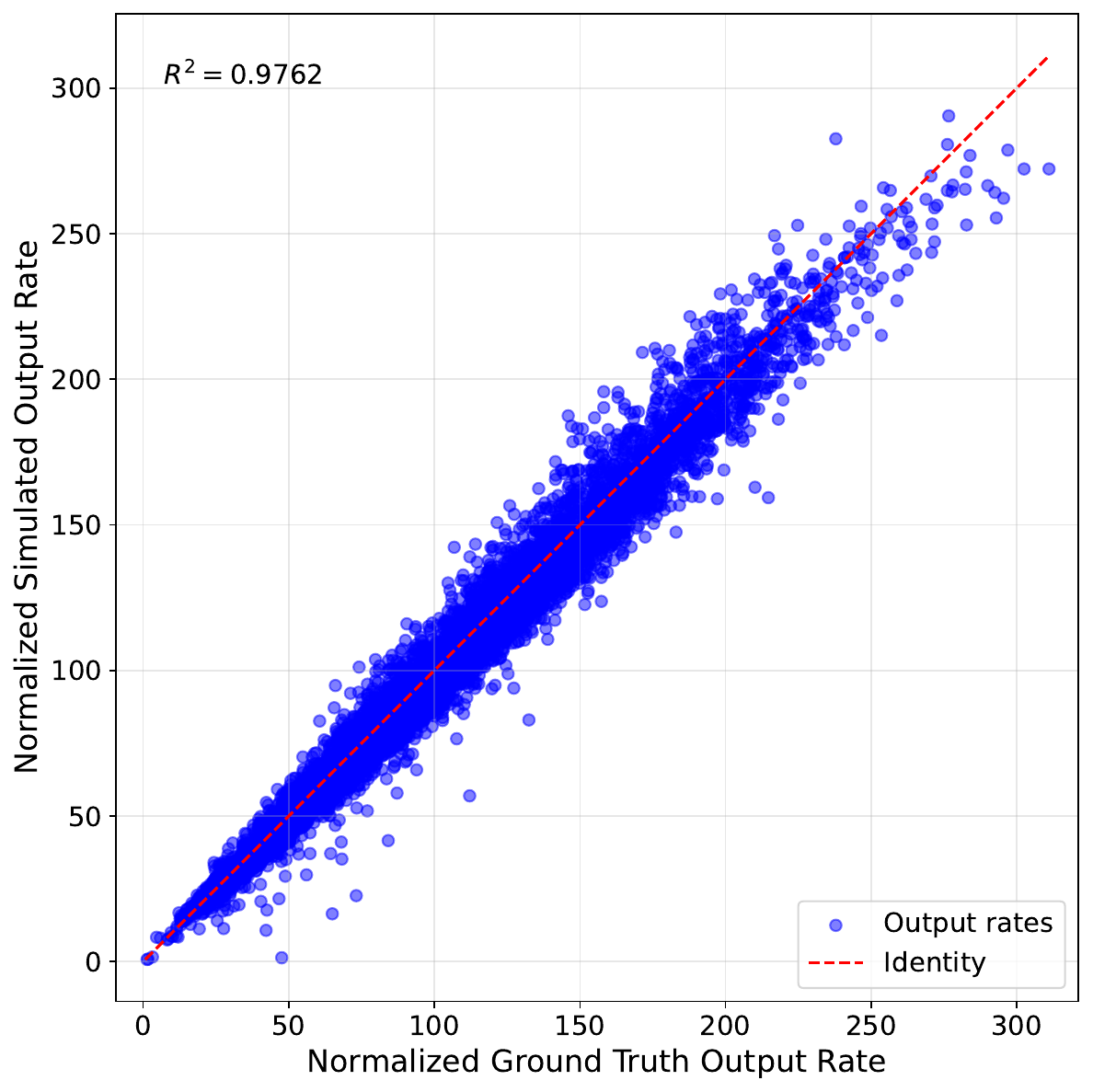}
\end{center}
\caption{Predicted versus ground-truth output rates for the hand-crafted simulator. The data points cluster along the identity line (in red), which indicates that the simulator has been well-calibrated with no systematic over- or underprediction. $R^2 = 0.976$ confirms that the simulator captures almost all of the variance in the ground-truth output.}
\label{fig:output_rates}
\end{figure}

\vspace{-3mm}

\begin{table}[thb]
\centering
\caption{The weighted average prediction errors of the learned simulator and the hand-crafted simulator. The hand-crafted simulator predicts the total throughput more accurately, whereas the learned simulator is superior at modeling the buffer states.}
\vspace{0.5em}  %
\label{tab:simulator_wapes}
\begin{tabular}{lcc}
\hline
\textbf{Metric} & \textbf{Hand-crafted Simulator} & \textbf{Learned Simulator} \\
\hline
Stage 1 Throughput & 12.0\% & 20.0\% \\
Stage 2 Throughput & 12.1\% & 18.5\% \\
\textbf{Stage 3 (Total) Throughput} & \textbf{9.5\%} & 15.8\% \\
Buffer State 1 & 18.0\% & 7.2\% \\
Buffer State 2 & 62.7\% & 25.6\% \\
Buffer State 3 & 16.5\% & 6.7\% \\
Buffer State 4 & 35.0\% & 8.9\% \\
\hline
\end{tabular}
\end{table}

We evaluate approaches using calibrated internal simulators that model system dynamics, including buffer behavior, inter-line interactions, and stage-specific processing characteristics. We compare our learned policies against historical human decisions replayed in the same simulated environments. We use offline RL and LLMs at different levels of state abstraction. We have a matched simulation environment for each method. For the offline RL, we evaluate the methods on two learned simulators, using the average of the simulators, whereas for LLMs, we use a hand-crafted simulator. %
Consequently, the two approaches should be evaluated on their own merits rather than compared against each other. For all results, we report 95\% confidence intervals. Due to computational constraints, this corresponds to evaluation variance. Each reported model was trained once, which means that true uncertainty including training variance may be wider.

The weighted absolute percentage errors (WAPE) against the ground-truth test set for one of the learned simulators and the hand-crafted simulator are shown in Table~\ref{tab:simulator_wapes}. The hand-crafted simulator measures total throughput more accurately, with a WAPE of 9.5\%. The WAPE for the learned simulator is 15.8\%. On the other hand, we see that the learned simulator is generally more accurate at buffer state prediction. Buffer State 2, showing the highest prediction error, measures the number of fully completed stage 2 outputs before they are consumed by stage 3, and its value should typically be minimized. Hence, the denominator for the WAPE calculation is very small, which artificially inflates the error number. The predicted versus ground-truth output rates for the hand-crafted simulator are plotted in Figure~\ref{fig:output_rates}, which indicate that the simulator has no systematic over- or underestimation bias. 

\section{Offline Reinforcement Learning}

\begin{figure}[t]
\begin{center}
\includegraphics[width=0.7\textwidth, trim={4mm 4mm 120mm 4mm}, clip]{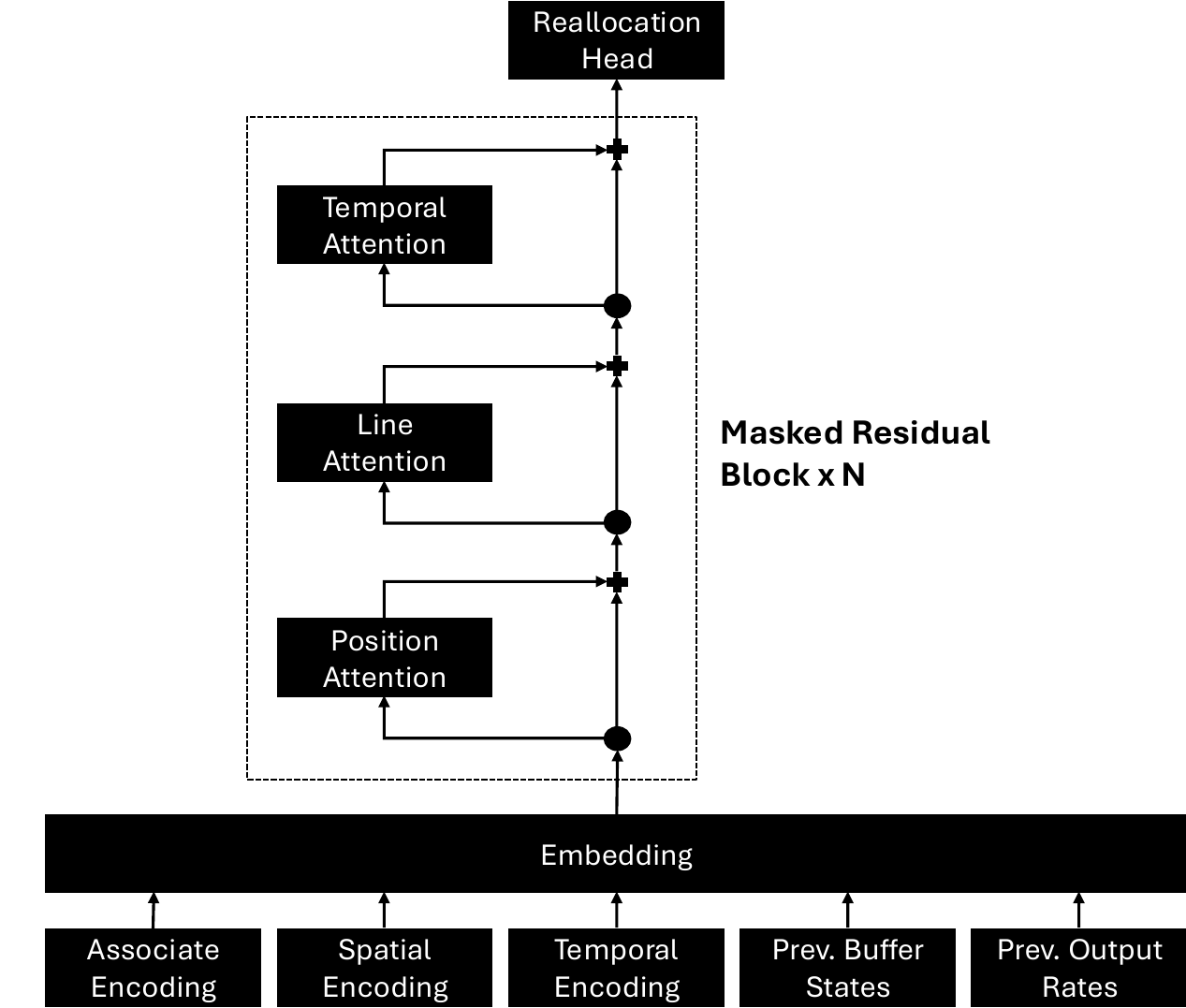}
\end{center}
\caption{The TGNN architecture consists of repeated masked residual blocks. Each residual block integrates information along the stages, then between the lines, and finally across the time dimension.}
\end{figure}

\vspace{-3mm}

Training policies for this domain is challenging. The state space consists of the current worker assignments, the throughput rates and the buffer states across all lines and stages, which means that the observation is very high-dimensional with complex spatiotemporal dependencies. We process raw inputs through embedding layers and MLPs to produce a feature tensor of shape $B \times T \times L \times P \times D$, where $B$ is the batch size, $T$ the number of timesteps in the episode, $L$ the number of lines, and $P$ is the number of position tokens per line. Each line consists of multiple staffable positions across stages, and we additionally concatenate buffer state features that capture inter-stage queue levels. $D$ is the feature dimension. Directly applying spatial-temporal transformers that attend jointly over all positions and timesteps (with causal masking) leads to quadratic complexity, which requires multi-node, multi-day training runs. To enable tractable experimentation, we developed a Transformer-GNN (TGNN) architecture inspired by graph multi-attention networks for traffic prediction \citep{zheng2020gman}. Rather than attending over all position-time pairs simultaneously, TGNN first applies graph attention over the spatial dimensions (positions and lines), and then applies temporal attention over timesteps. This factorization reduces computational complexity substantially, with all models fitting on a single GPU and training time being reduced by 95\% without a loss in performance.

The action space is another challenge. At any point in time, the manager can reassign any subset of workers to any valid positions. The joint action space grows combinatorially, and a naive enumeration of the action space is intractable. We address this through factorization. Instead of performing a joint assignment, each worker independently decides whether to stay or move, and if moving, selects a destination. The policy network implements this using a reallocation head that operates on the position features that are produced by the TGNN backbone. Inspired by the attention mechanism, we project position features to queries and keys, and then compute the dot-product similarities between all position pairs. For each occupied position, the softmax over these similarities represents a probability distribution over destinations. This also includes the current position, which represents no movement. We apply a threshold: workers only move if the probability of staying is below 0.1, which ensures that only a limited number of reallocations is performed due to their cost. Thanks to this factorization, made possible by the conditional independence assumption over worker movements (a simplification), we can avoid exponential enumeration of joint assignments, and we get polynomial time complexity. The movement threshold limits simultaneous reassignments, which mitigates the practical impact of this assumption.

We train policies from historical data using offline reinforcement learning. Online reinforcement learning is not suitable, as it would disrupt live operations. With very large action spaces, value-based methods that learn $Q(s, a)$ for all actions are problematic. Rarely-seen actions receive overoptimistic value estimates, which can significantly distort the policy learning \citep{kumar2020conservative, mathieu2023alphastar}. Instead, we adopt an actor-critic approach with a state-value critic $V(s)$. Advantages are computed as $A(a_t, s_t) = G_t - V(s_t)$. We compared against GAE \citep{schulman2015high}, TD(0) and V-Trace \citep{espeholt2018impala}, and MC-returns for calculating the returns $G_t$, and did not find significant differences between the methods, so we opted for MC-returns for simplicity. Then, we add behavioral regularization following \citet{fujimoto2021minimalist} to keep the policy close to demonstrated behavior. The policy gradient objective is: 

\begin{equation} \mathcal{L} = -\sum_{t} \text{sg}[A(a_t, s_t)] \log \pi(a_t | s_t) - \alpha \sum_{t} \log \pi(a_t | s_t), \end{equation} 
where $A(a_t, s_t)$ is the advantage function, $\text{sg}[\cdot]$ denotes stop-gradient and $\alpha$ controls the strength of the regularization. The first term corresponds to advantage-weighted regression, and the second term encourages the policy to stay close to the demonstrated behavior, to prevent the exploitation of value function errors.

As a simpler alternative to offline actor-critic, we evaluate fine-tuned behavior cloning (BC-FT) \citep{mathieu2023alphastar}. First, we train a policy via supervised learning on all available demonstrations, and then we fine-tune the policy on the subset of trajectories with the highest cumulative reward. Offline RL can theoretically exceed the performance of the demonstrator by learning from suboptimal actions, but BC-FT is limited by the demonstration quality. However, we found BC-FT to have practical benefits. It is simpler to implement, its training dynamics are more stable and it is less sensitive to hyperparameters. We use identical network architectures for both approaches. For the value function, we simply replace the worker reallocation head with an MLP head.

Table~\ref{tab:offline-rl} presents our main results. We evaluate our trained policies on held-out shifts by initializing the simulator to historical states and rolling out learned policies. We compare the resulting throughput against replayed human decisions in the same simulator. Both offline RL approaches outperform the behavior cloning baseline, which imitates all historical actions without distinguishing their quality. Offline actor-critic improves 2.4\% over the human decisions, and BC-FT achieves a 2.1\% improvement. The ranking of the methods was consistent across both learned simulators, which also suggests that the results are robust across different simulators. All learned methods outperform replayed human decisions. This does not imply that learned policies exceed human performance in reality. Instead, when historical human actions are replayed in a simulator, we get a mismatch if the simulator diverges from reality. Learned policies adapt to the simulator states, but replayed decisions cannot. 

Overall, the performance of BC-FT suggests that if a sufficient quantity of high-quality human demonstrations exists, sophisticated offline RL may offer limited additional benefit. We also observed that BC-FT was more robust during training. Offline actor-critic required early stopping to prevent overfitting, whereas BC-FT was more stable. In preliminary analyses, BC-FT policies outperformed offline actor-critic policies when they were evaluated on states outside the training distribution. These practical advantages mean that BC-FT is a more attractive starting point for operational deployment, especially given that the deployment goal is to achieve performance parity across facilities and support human decision-making rather than exceed the best human operators. By learning from high-quality demonstrations, BC-FT can elevate all sites to top-performer levels without the need to extrapolate beyond observed behavior.

\begin{table}[thb]
\centering
\caption{Offline RL results in two internal learned simulators. Throughput improvement (percentage) is measured relative to replayed human decisions. Results are averaged over two different simulators to prevent overfitting to simulator-specific dynamics.}
\label{tab:offline-rl}
\vspace{0.5em}  %
\begin{tabular}{lcc}
\hline
\textbf{Method} & \textbf{Throughput Improvement} & \textbf{95\% CI} \\
\hline
Offline Actor-Critic & \textbf{+2.4\%} & [+2.3\%, +2.6\%] \\
Fine-Tuned Behavior Cloning & +2.1\% & [+2.0\%, +2.3\%] \\
Behavior Cloning & +1.6\% & [+1.4\%, +1.8\%] \\
\hline
Human Decisions (replayed) & 0\% & -- \\
\hline
\end{tabular}
\end{table}

\section{Large Language Models}

While the offline reinforcement learning method operated on highly detailed state descriptions, this representation does not fully mirror how managers actually reason about operations. Managers typically monitor summary statistics like throughput rates, buffer levels, and staffing counts rather than raw sensor data. LLMs offer potential advantages for this setting. They can work with flexible text-based inputs, they are capable of commonsense reasoning and they support natural language interfaces, which could eventually be useful for explaining the decisions or having a dialogue with human operators. However, it is unclear if LLMs can translate this high-level reasoning ability into effective strategic decision-making. We systematically evaluate how much task-specific adaptation is required when LLMs are given abstracted, human-readable state descriptions. 

To match the abstraction level at which LLMs operate, we developed and calibrated a hand-crafted simulator that models the item flow through the system using explicit causal rules. This simulator tracks aggregate process rates and buffer states, rather than individual position dynamics, which mirrors the summary statistics that are presented to the LLM. The hand-crafted simulator is also faster to run, which is a practical advantage for LLM evaluation, and it has causal structure, which means that interventions have interpretable effects. However, the offline RL policy operates on position-level features, which are outside the scope of this simulator due to the difficulty of simulating these with pre-defined rules. This makes direct comparison of LLM and offline RL methods infeasible. Despite this limitation, we validated the accuracy of the hand-crafted simulator against ground-truth data, and we found that the throughput prediction accuracy exceeded that of the learned simulators used for the offline reinforcement learning experiments. 

\begin{figure}[t]
\begin{center}
\includegraphics[width=0.7\textwidth]{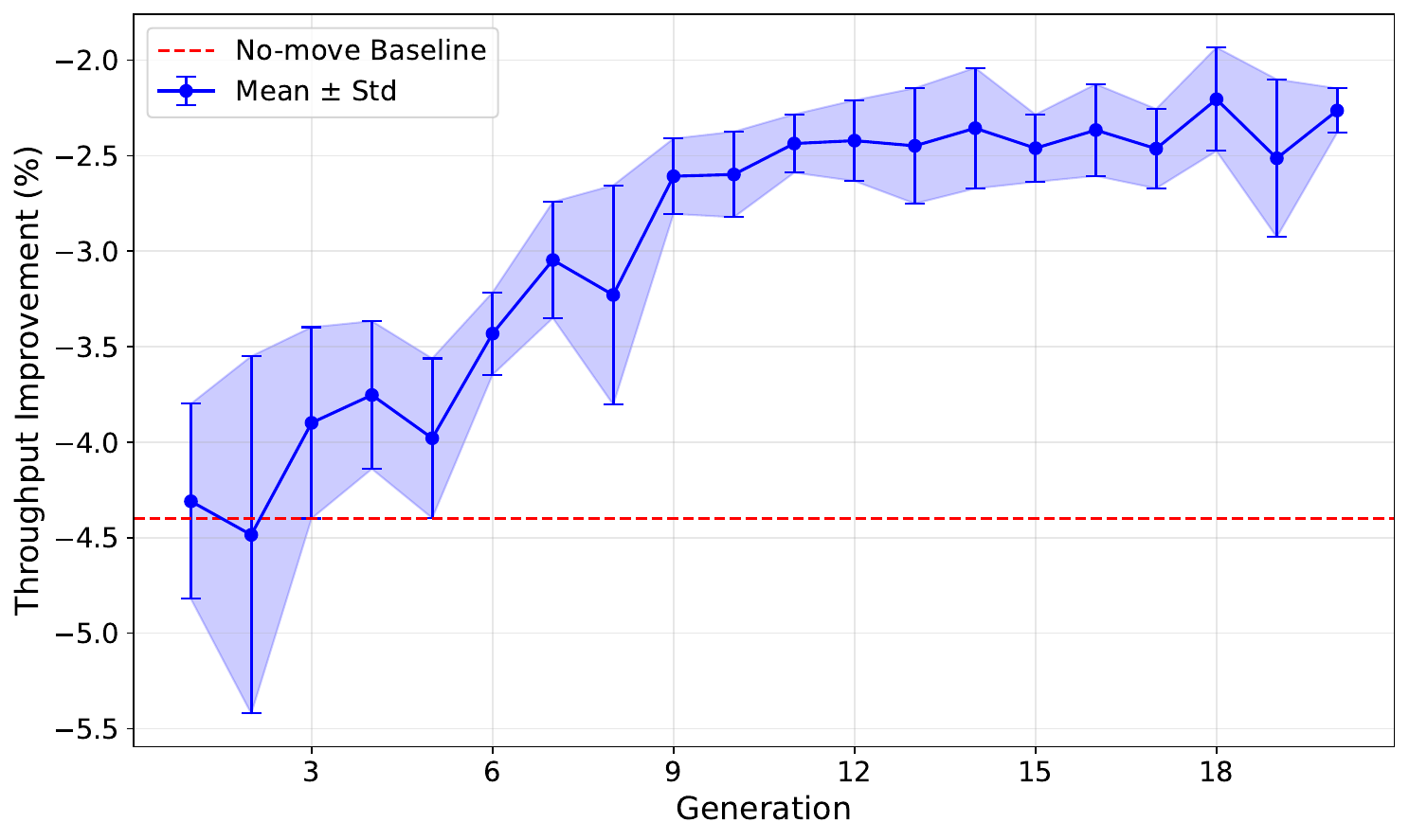}
\end{center}
\caption{The learning curve for \textbf{O}ptimization by \textbf{PRO}mpting (OPRO). We see that the performance improves steadily before convergence.}
\end{figure}

\begin{table}[t]
\centering
\caption{Prompting and prompt optimization results in the hand-crafted simulator. Throughput improvement is measured relative to replayed human decisions.}
\vspace{0.5em}  %
\label{tab:prompting}
\begin{tabular}{lcc}
\hline
\textbf{Method} & \textbf{Throughput Improvement} & \textbf{95\% CI} \\
\hline
\multicolumn{3}{l}{\textit{Automatic Prompt Optimization}} \\
OPRO + Self-Refine & -2.0\% & [-2.3\%, -1.8\%] \\
Evolutionary Reflexion & -2.0\% & [-2.3\%, -1.7\%] \\
Meta Agent Search & -2.7\% & [-3.0\%, -2.4\%]\\
\hline
\multicolumn{3}{l}{\textit{Hand-Crafted Prompting}} \\
Self-Refine & -2.1\% & [-2.4\%, -1.8\%] \\
Self-Consistency & -2.5\% & [-2.7\%, -2.2\%]\\
Chain-of-Thought & -2.7\% & [-2.9\%, -2.4\%] \\
Default Prompting & -3.5\% & [-3.8\%, -3.3\%]\\
\hline
Human Decisions (replayed) & 0\% & -- \\
No Reallocation Baseline & -4.6\% & -- \\
\hline
\end{tabular}
\end{table}

We represent system state as structured text that describes each line's staffing, throughput rates, and buffer status. The LLM receives this description and a task description, and it has to output a JSON-formatted list of worker reassignments. We evaluated several prompting strategies: default prompting with task instructions and domain knowledge, chain-of-thought prompting for step-by-step reasoning \citep{wei2022chain}, also with few-shot examples, self-consistency that samples multiple responses and selects the most common action \citep{wang2022self}, and self-refine, which iteratively generates, critiques, and improves the proposed actions \citep{madaan2023self}. Notably, including few-shot examples actually degraded performance because the models copied specific movements from the examples rather than reasoning thoroughly about the current state. We also explored automatic prompt optimization methods. OPRO \citep{yang2023large} uses an LLM to iteratively improve prompts based on performance feedback. We combined OPRO with Self-Refine, using OPRO to optimize the Self-Refine prompts. We used Reflexion \citep{shinn2023reflexion} with an additional evolutionary long-term memory of successful strategies. We call this method Evolutionary Reflexion. Finally, we also evaluated Meta Agent Search \citep{hu2024automated}, which uses a meta-agent to generate agent architectures. 

Table~\ref{tab:prompting} shows the results using Qwen2.5-32B-Instruct \citep{qwen2025qwen25technicalreport}. All methods outperform the no-reallocation baseline, which confirms that LLMs can generate useful recommendations. Self-Refine was the best among the hand-crafted prompts, and its throughput was 2.1\% below the replayed human decisions. OPRO was slightly better, but overall, the performance of the hand-crafted prompt was roughly similar to the automatically learned one. Notably, Meta Agent Search underperformed the simpler methods, which we attribute to the relative complexity of the method, which was too much for the 32B Qwen model to handle. These results indicate that prompting alone is insufficient for approaching human decision-making on this task. We attribute this to the complexity of the task. Because of the highly non-linear dynamics, the mapping from states to actions is too complex to be specified in a prompt, and must be learned from experience instead.

To close the gap, we developed a fine-tuning pipeline that combines supervised learning on historical data with preference optimization on synthetic data. For supervised fine-tuning, we converted the behavior cloning dataset into an instruction-tuning format, and trained the model to predict the staffing decisions made by the humans. After this, the performance was within 0.5\% of replayed human decisions, which is a large performance improvement over prompting but still below humans. In our analyses, we found that a key limitation of SFT, and especially rejection-sampling-based fine-tuning (RFT) \citep{yuan2023scalingrelationshiplearningmathematical}, is that it cannot teach the model what \emph{not} to do. In our preliminary experiments with RFT, we generated training data from states where the chosen action improved throughput. This biased the dataset toward reallocations. Because of this, the model recommended excessive movements and failed to learn that maintaining current staffing is often optimal.

Direct Preference Optimization (DPO) \citep{rafailov2023direct} tackles this limitation by learning from action comparisons. We generated synthetic preference data by sampling pairs of actions for each training state, and evaluated them via short simulator rollouts. The action that achieves the better throughput becomes the preferred response. This allows the model to learn that maintaining the current staffing often outperforms unnecessary movements. Additionally, we found that standard DPO suffers from likelihood displacement in our setting, where the model decreases the probability for both preferred and non-preferred responses \citep{razin2024unintentional}. Adding a supervised loss term on the preferred actions made the training more stable and increased the gap between preferred and non-preferred action likelihoods, as expected. Iterative DPO, where we regenerate preference data using the updated model and repeat fine-tuning, also proved beneficial \citep{pang2024iterative}. Each iteration simulates one batch of manager feedback, where the model proposes actions, receives evaluations from the simulated manager and refines its policy accordingly. 

Table~\ref{tab:llm} shows the results. We see clear improvements from performing fine-tuning, even though we used a smaller model (Qwen2.5-14B-Instruct) due to computational constraints. Given models of the same size, we would expect the advantage fine-tuning has over prompting-based approaches to be even greater, but prompting results might also improve with larger models. SFT approaches human performance, whereas DPO without SFT matches human performance after two iterations. This shows that synthetic preferences can generate competitive policies. The most effective combination is first performing SFT on the historical human decisions, and then using iterative DPO to refine the decision quality using the feedback from the simulator, and this approach outperforms the replayed human decisions by 0.6\%. This iterative process, where we sample from the current policy, collect preferences and update the policy, mirrors the core loop of reinforcement learning \citep{xiong2023iterative}, which also explains why further iterations continue to improve performance. 

\begin{table}[thb]
\centering
\caption{LLM results in hand-crafted simulator using Qwen2.5-14B-Instruct as the underlying model. Throughput improvement is measured relative to replayed human decisions.}
\vspace{0.5em}  %
\label{tab:llm}
\begin{tabular}{lcc}
\hline
\textbf{Method} & \textbf{Throughput Improvement} & \textbf{95\% CI} \\
\hline
SFT + DPO (2 iterations) & \textbf{+0.6\%} & [+0.5\%, +0.7\%]\\
SFT + DPO (1 iteration) & +0.3\% & [+0.1\%, +0.4\%] \\
DPO only (2 iterations) & +0.0\% & [-0.2\%, +0.2\%] \\
DPO only (1 iteration) & -0.4\% & [-0.7\%, -0.2\%]\\
SFT only & -0.5\% & [-0.6\%, -0.4\%] \\
Self-Refine (Qwen 32B) & -2.1\% & [-2.4\%, -1.8\%] \\
\hline
Human Decisions (replayed) & 0\% & -- \\
\hline
\end{tabular}
\end{table}

Overall, the experiments demonstrate that prompting mid-sized open-weight language models like Qwen2.5-32B-Instruct is insufficient for complex strategic decision-making in operational control. However, incorporating historical decision data helps, yielding greater gains when combined with DPO than when learning from synthetic preference data alone. Finally, given that further iterations of DPO continue to improve performance, this suggests that continued learning from real manager feedback after deployment could give further gains.

\section{Discussion}

Both offline RL and fine-tuned LLMs are viable approaches for AI-assisted operational decision-making. They generate recommendations that managers can modify and override based on knowledge that the AI system lacks. In both settings, we saw imitation-based methods achieve strong performance in this task: BC-FT came within 0.3\% of offline actor-critic, and SFT alone brought LLMs within 0.5\% of replayed human decisions before preference optimization. Offline RL is well-suited when detailed state information exists and fast inference is a priority. LLMs can be considered when interpretability and natural language interaction with the operators are priorities. In our experiments, we found prompting alone to be insufficient, despite doing our best to optimize the prompt. We believe that effective staffing decisions may depend on patterns that are learned through experience and are hard to formalize. Hence, designing mechanisms to elicit this knowledge naturally is an essential direction for future work. 

\textbf{Limitations.} Our formulation assumes that the system dynamics are stationary, and it does not capture all factors present in real operations. Some portion of the offline RL improvement over replayed human decisions may be due to the policy being able to adapt to evolving simulator states whereas historical decisions cannot respond when simulated conditions diverge from the historical trajectory. However, the ranking of methods was consistent across both learned simulators. Our DPO experiments used synthetic preferences from the same simulator that was used for evaluation, serving as a controlled proxy for real manager feedback to validate the iterative framework. While training and evaluation states are fully disjoint, they both rely on the same dynamics function, which means that policies may learn simulator-specific patterns that do not fully transfer to real operations. Furthermore, real manager preferences reflect factors other than throughput, which the simulator does not capture. When comparing prompting and fine-tuning, the different model sizes confound the comparison.

\textbf{Future Work.} Real-world deployment would require addressing additional constraints such as distribution shift across facilities and real-time latency. In particular, our iterative DPO framework was designed with this in mind. Replacing the simulator-generated preferences with real manager feedback requires no changes to the core method. This could be done by showing managers two actions and asking which they prefer, or letting them propose a third alternative if neither recommendation is satisfactory and to explain their reasoning. This can also capture domain knowledge that is currently outside the scope of the simulator, such as operational priorities and real-time exceptions. An online learning setting where the system collects feedback and periodically updates itself is a promising direction. Focusing on improving the natural language justifications for the recommendations could support manager trust and make feedback more natural to provide. On the technical side, potential directions for model improvement include scaling up the model size to larger LLMs than those included in this study, giving the LLMs tool access for data analysis, and performing reinforcement learning-based fine-tuning of the LLMs.

\bibliography{iclr2026_conference}
\bibliographystyle{iclr2026_conference}

\end{document}